# Comment on "Modeling rapid language learning by distilling Bayesian priors into artificial neural networks"

Orr Well[1], Idan Tarshish[1], Nur Lan[2]*, Roni Katzir[1]*

[1]Tel Aviv University

[2]École Normale Supérieure

**McCoy & Griffiths[1] (henceforth M&G) suggest that a Bayesian prior can be distilled into Artificial Neural Networks (ANNs) through Model-Agnostic Meta-Learning (MAML)[2]. They support this empirically by showing that meta-trained networks demonstrate formal language learning abilities comparable to Yang & Piantadosi's Bayesian learner[3] (henceforth Y&P), significantly outperforming standard ANNs. We point out that under the standard interpretation of a prior, M&G's procedure does not actually instill one; it merely initializes network weights favorably, leaving the objective function unchanged. We then consider a more permissive interpretation, where the system as a whole can be seen as implementing a Bayesian learner even without an explicit prior in the objective. We show that this interpretation faces nontrivial challenges. Finally, we assess how well MAML approximates the empirical results of Bayesian learning, showing that unlike genuine Bayesian learners, M&G's model overfits and generalizes poorly to unseen data.**

*These authors jointly supervised this work.

## M&G's methodology

M&G's distillation method proceeds as follows. First, a target bias is defined using a symbolic Bayesian model which, following Y&P, probabilistically combines formal primitives to create definitions of formal languages. This model's prior exhibits a simplicity bias, assigning higher probabilities to definitions that use fewer primitives. Second, a Long Short-Term Memory (LSTM) "student model" undergoes meta-training: in each episode, a copy of the LSTM is trained on a formal language sampled from the Bayesian model's prior, and the original LSTM parameters are updated based on the copy's loss on a held-out test set. After 25,000 episodes, the checkpoint with the lowest loss on a validation meta-set of 500 formal languages is selected.

M&G's hope is that exposure to many formal languages sampled from the Bayesian prior will somehow transfer that prior to the LSTM. This claim is significantly bolder than MAML's original formulation as merely a method for finding an initialization that enables strong performance on new tasks to be achieved with relatively few gradient steps.[2] MAML may indeed be an effective initialization method; what we argue against is M&G's novel claim that MAML distills a Bayesian prior into the LSTM.

## The role of priors

We start by making explicit a terminological point that is implicit in M&G's paper. On a common understanding, a Bayesian prior operates at the level of the objective function: the prior probability is jointly optimized with data likelihood, thereby actively regularizing learning and, when appropriately specified, enabling generalization from limited data. MAML, by contrast, only shifts the initial hypothesis to a more favorable region of the search space (see ref. [4] for discussion of what M&G's initialization may encode). This leaves the objective function untouched: M&G adapt their meta-trained networks to new tasks by minimizing standard CE loss, with no prior entering the objective.

Therefore, MAML clearly does not distill a Bayesian prior in the usual sense — that is, a distribution over hypotheses (e.g., possible model parameters), not a single hypothesis such as an initial parameter setting. Still, one may wonder how much this matters: even if MAML only serves to set the initial hypothesis rather than introduce a prior into the objective, perhaps the system as a whole behaves as if it were Bayesian. Indeed, it has been argued that meta-trained network parameters can be interpreted as estimating a prior distribution.[5] As we show below, however, this interpretation is of questionable relevance to M&G's setting. This still leaves open the possibility that M&G's framework, while not Bayesian even as a whole, is an empirical approximation of a Bayesian learner. As we show, however, M&G's model provides a much weaker approximation than suggested by their own figures.

**The role of early stopping**

The objective function M&G use, namely CE, has been shown to be inadequate for formal language learning, favoring incorrect solutions over correct ones.[6–8] Given this, the best M&G can hope for is to shift the initial model weights toward a good-enough solution and rely on subsequent training to remain in that region. ANN optimization is limited in ways that are only partially understood (in particular, it is not guaranteed to converge to a global minimum), which may prevent a flawed objective from doing too much harm to a strong initialization. However, prior work indicates that standard training is not so weak as to mitigate the issues with CE as an objective.[6,7]

Inspection of M&G’s code reveals a further constraint on the search: language-specific training is limited to just 6-15 epochs, depending on the training size. We show that this is not a technicality: without early stopping, M&G’s performance deteriorates as training proceeds. For each of M&G’s 56 test formal languages, we compute the CE loss per epoch on a test set of one million strings. Figure 1 demonstrates an overfitting pattern: as training loss decreases, test loss gradually increases, most clearly for languages where unseen strings constitute a larger proportion of the test set (see the supplement for full results and further discussion). Bayesian learners normally resist overfitting and converge on correct solutions regardless of training duration; M&G’s models do not. This highlights that their apparent success is more plausibly due to favorable initialization combined with severely limited training, rather than to any Bayesian priors entering the objective.

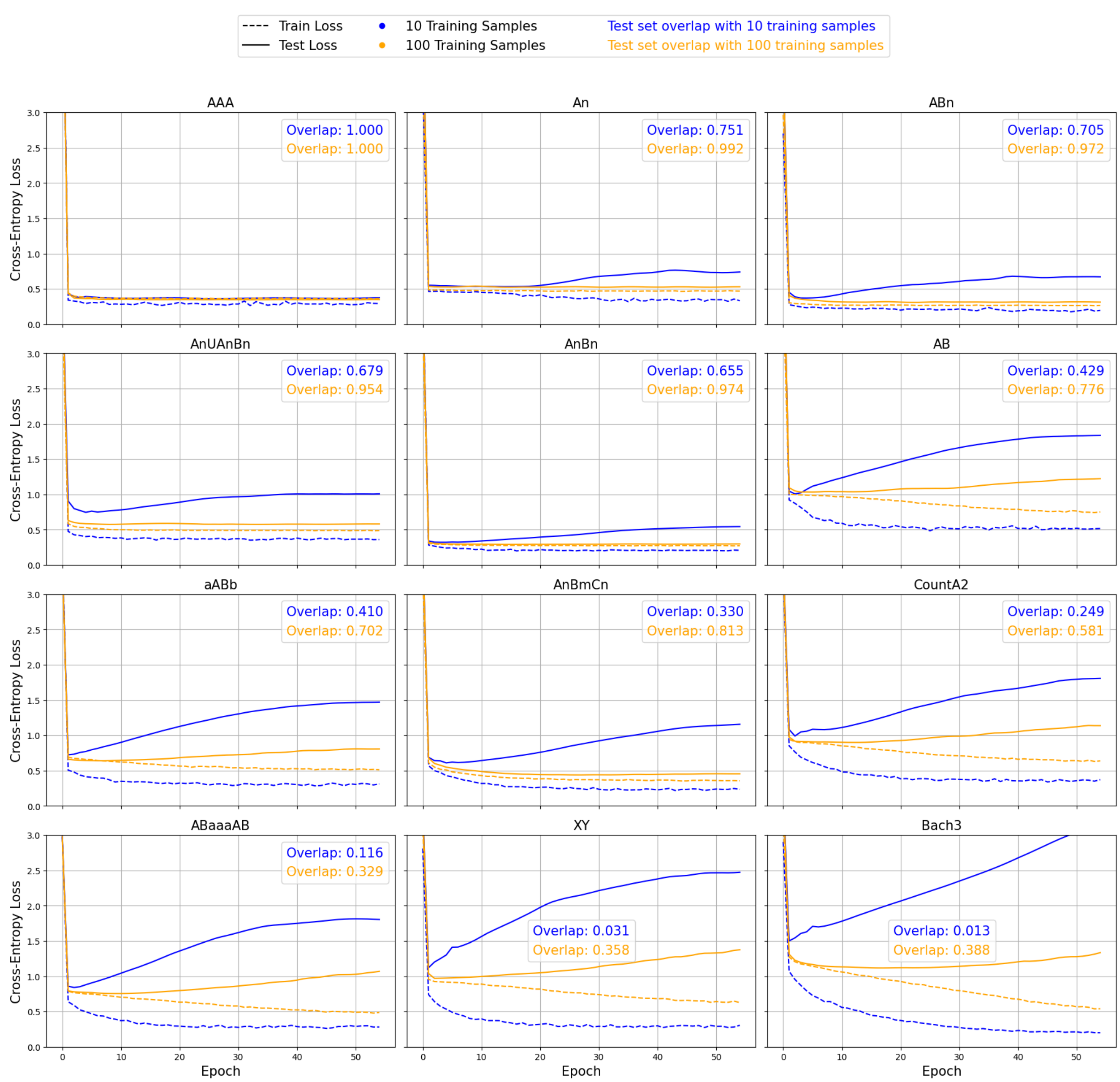

**Figure 1**. Training and test CE (test set size = 1M) across epochs for selected formal languages. When the training-test data overlap is small, test performance gradually deteriorates, indicating overfitting. See Appendix A for language definitions.

**Early-stopping as a prior?**

Could early-stopping be a feature rather than a bug? One might argue that combining a shifted initialization with early-stopping effectively induces a prior: hypotheses closer to the initialization in parameter space are reachable with fewer gradient steps, and thus implicitly assigned higher prior probability. Learning would then trade off loss minimization against staying close to the initialization, using early-stopping as a form of regularization. This perspective is suggested by Grant et al.[5], whom M&G cite in support of their claim that MAML estimates a Bayesian prior.

This view is problematic for several reasons. First, Grant et al.'s theoretical results were only shown to hold under specific assumptions — linear regression with L2 regularization and a Gaussian prior over weights — which are not satisfied in M&G's setting. Second, even if those results extended to formal language learning, there is no reason to expect the prior induced by the initialization to match M&G's original Bayesian model's prior (see also ref. [4]). Third, even granting that the initialization encodes an approximation of the desired prior, it remains unclear how M&G's implementation, with a fixed number of epochs, could realize the trade-off between loss and proximity to the initialization described by Grant et al. Rather, for any predetermined k, one obtains at best a distribution over hypotheses reachable after exactly k iterations of a stochastic optimizer. The value of k is not something MAML derives, and unless it happens to be chosen with exceptional luck, the correct solution is unlikely to fall within the restricted set of reachable hypotheses.

**Empirical consequences**

Nonetheless, M&G show that their model performs similarly to Y&P's Bayesian model when evaluated on the same formal languages. In light of the problems mentioned above, how can we explain this similarity? We argue that it is merely an artifact of another methodological flaw: the choice of success criterion. M&G's evaluation relies on an F1-score that only considers the 25 most frequent strings in the language. Since the CE objective biases models toward high-frequency patterns, this metric artificially inflates performance and rewards overfitting. In particular, it does not reliably measure generalization to unseen string lengths (meta-learning indeed tends to fail on longer strings; see ref. [4,9]).

As detailed in the supplement, more appropriate metrics reveal substantial generalization gaps between M&G's and Y&P's models: in cases where Y&P's model perfectly learns the underlying grammar, M&G's model performs poorly on longer strings, even with substantial training data. We demonstrate this using two of M&G's test languages, AnBn and AnBnCn, which can be recognized using a counting mechanism and offer a particularly clear generalization criterion: generalization to unseen n values. As shown in Figure 2 for AnBn, beyond a certain n M&G's model fails to predict the end-of-sequence (EOS) token's timing, indicating that it has not generalized to the target counting pattern. This failure is likely not due to architectural limitations, since counters are provably representable by LSTMs.[6,10] While the standard for a good approximation may depend on the task at hand, this suggests that M&G's approximation of Bayesian generalization is poorer than implied by their own work.

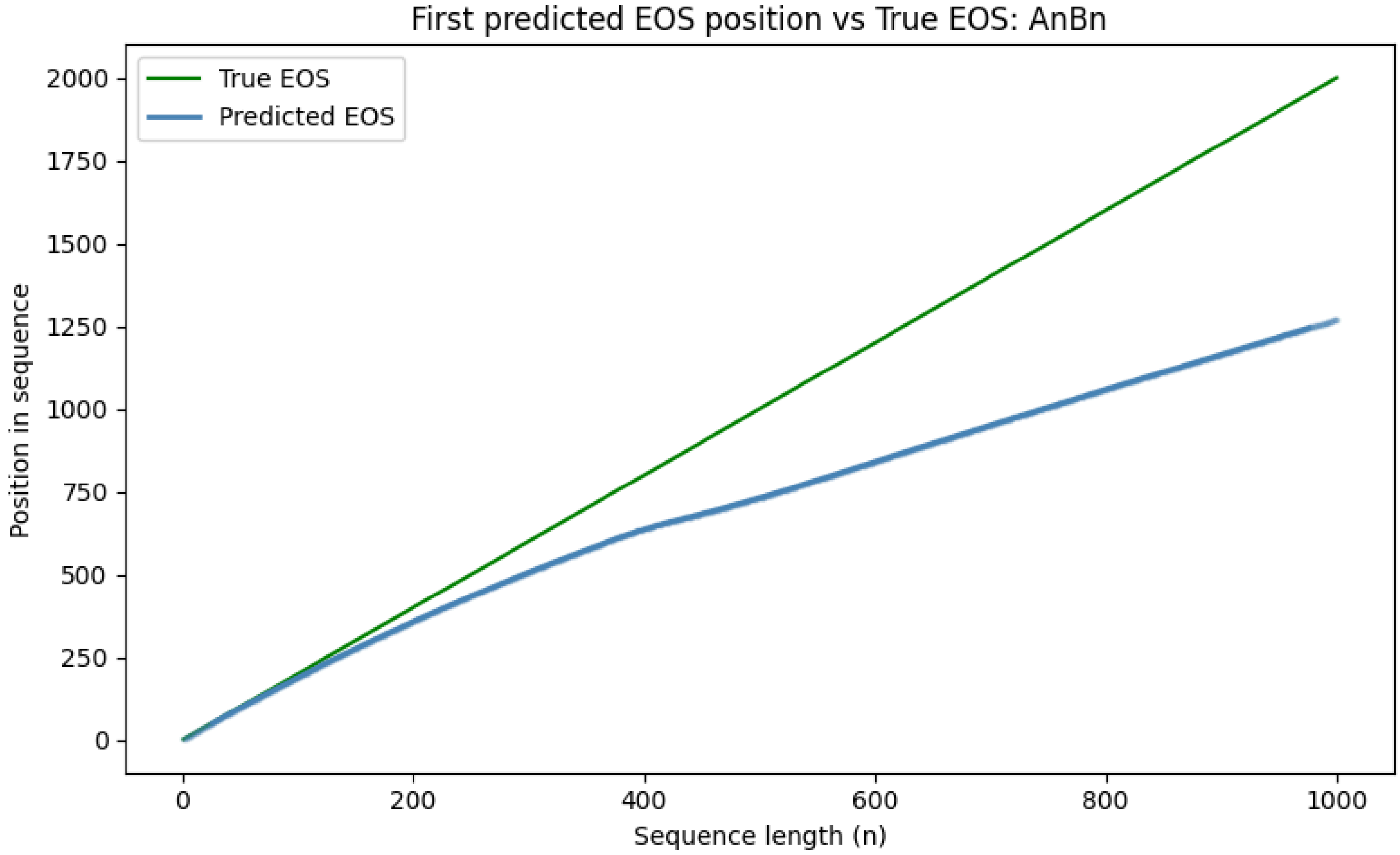

**Figure 2**. Predicted vs. ground-truth EOS position in AnBn strings (train set size = 1K). M&G's model incorrectly predicts the EOS increasingly earlier as n grows.

**Acknowledgements**

We thank R. Thomas McCoy and Thomas L. Griffiths for valuable feedback on an earlier version of this manuscript. RK acknowledges support from ISF grant #1083/23, the Alexander von Humboldt Foundation, and the NOMIS Foundation. This project was provided with computer and storage resources by GENCI at IDRIS thanks to the grant AD011013783R3 on the supercomputer Jean Zay's V100 and CSL partitions.

# Supplementary Information

Both experiments below start from the meta-trained model *meta_lm_hidden1024_39.weights*, a two-layer LSTM-based language model with a hidden size of 1024, selected at random among the 40 models published by M&G on OSF (which differ only in the random seed used during meta-training). Given the generality of our arguments, we don't expect results to differ meaningfully across M&G's other models. Unless otherwise stated, all hyperparameters follow M&G. Experiments were conducted using a single NVIDIA RTX A6000 GPU (48GB VRAM), 4 CPU cores, and 8GB of RAM per job.

## Experiment 1: early stopping

**Experimental setting**

This experiment builds on M&G's methodology by training their meta-trained model further on the 56 formal languages introduced by Y&P. The original framework is preserved, with the goal of assessing symbolic generalization from a limited set of just 10 or 100 training samples. The main technical modification lies in increasing the depth of optimization: we extend M&G's language-specific training phase (5 SGD epochs + 1 Adam epoch for a training set of size 10, 10 SGD epochs + 5 Adam epochs for a training set of size 100) to 5 SGD epochs followed by 50 Adam epochs (the specific details are arbitrary; as shown in "Optimizer variations" below, other choices yield similar results). Training and test CE are computed at the end of each training epoch, with test CE computed over a dataset of one million sample strings. For each language, we also measure the overlap between the training and test sets, defined as the proportion of test strings that appear at least once in the training set, for reasons explained below.

**Results**

The results are shown in Figure 3, with languages ordered by decreasing train-test overlap. As argued in the main text, longer training leads to overfitting, confirming that no meaningful prior is retained from the MAML process. We elaborate on this below, showing that while overfitting is more apparent in some languages than in others, it occurs consistently across all 56 languages.

Whether the model appears to generalize depends critically on the degree of overlap between training and test sets. Higher overlap correlates with lower final-epoch test CE, both for models trained on 10 samples (Pearson's $r = -0.75$) and 100 samples (Pearson's $r = -0.65$). Note that our overlap measure is conservative: a memorizing model can exploit not just identical strings but also shared substrings, inflating performance even on test strings unseen during training. A more robust measure would likely yield an even stronger correlation.

For permissive languages — those with relatively weak constraints on valid strings — the test CE clearly increases over time, indicating that the model fails to generalize to the underlying distribution and remains highly surprised by unseen samples. Notably, this pattern appears largely independent of computational complexity: it is observed in complex languages (e.g., the context-sensitive Bach3, XX, and WeW; see Appendix A for definitions) and simple ones alike (e.g., the regular languages AB, aABb and ABaaaAB). What these languages share is a large variety of valid strings, resulting in minimal train-test overlap.

For more restrictive languages, including finite ones such as AAA and AAAA, test loss remains close to training loss. This might initially suggest successful generalization but is better explained by the substantial train-test overlap, which makes memorization of the training data disproportionately beneficial for test performance. Some computationally complex languages fall into this category as well. For instance, the context-sensitive language A2en has a test set of only 10 unique strings, with nearly 96% of the one million test samples being repetitions of "a" and "aa", both present in the training set. By contrast, the regular language aABb has 13,717 unique test strings, only 6 of which overlap with the 10-sample training set. We thus have an explanation for the otherwise puzzling observation that performance appears stronger for some computationally complex languages: these languages tend to be more restrictive, resulting in higher overlap.

Similar reasons likely explain why divergence between the training and test curves is overall greater with 10 training samples compared to 100: the larger the training set, the higher the train-test overlap. Taken together, these findings argue against a Bayesian characterization of the model and instead indicate systematic overfitting.

**Optimizer variations**

The experiment above modifies M&G's two-stage optimization procedure by extending the Adam phase to 50 epochs. To assess whether this specific choice is what drives the observed overfitting behavior, we repeat the experiment with two alternative optimizer schedules: one that instead extends the SGD phase (55 SGD epochs + 1 Adam epoch), and another that extends both phases (23 SGD epochs + 23 Adam epochs). As shown in Figure 4, these variations do not lead to any significant differences in performance, confirming that the observed overfitting behavior is not sensitive to the specific optimizer schedule.

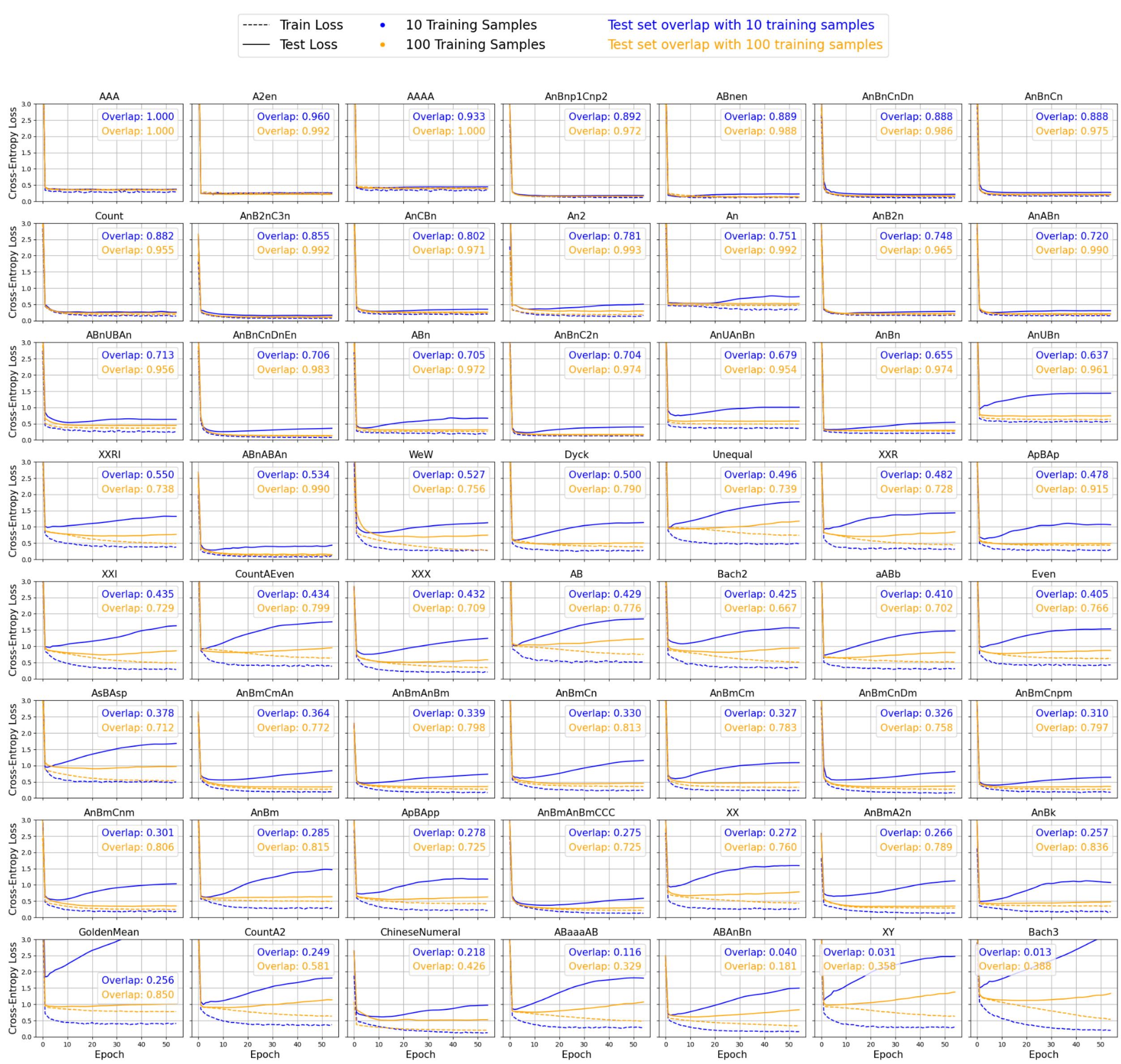

**Figure 3**. Training and test CE (test set size = 1M) across epochs for Y&P's 56 formal languages. When the training-test data overlap is small, test performance gradually deteriorates, indicating overfitting. See Appendix A for language definitions.

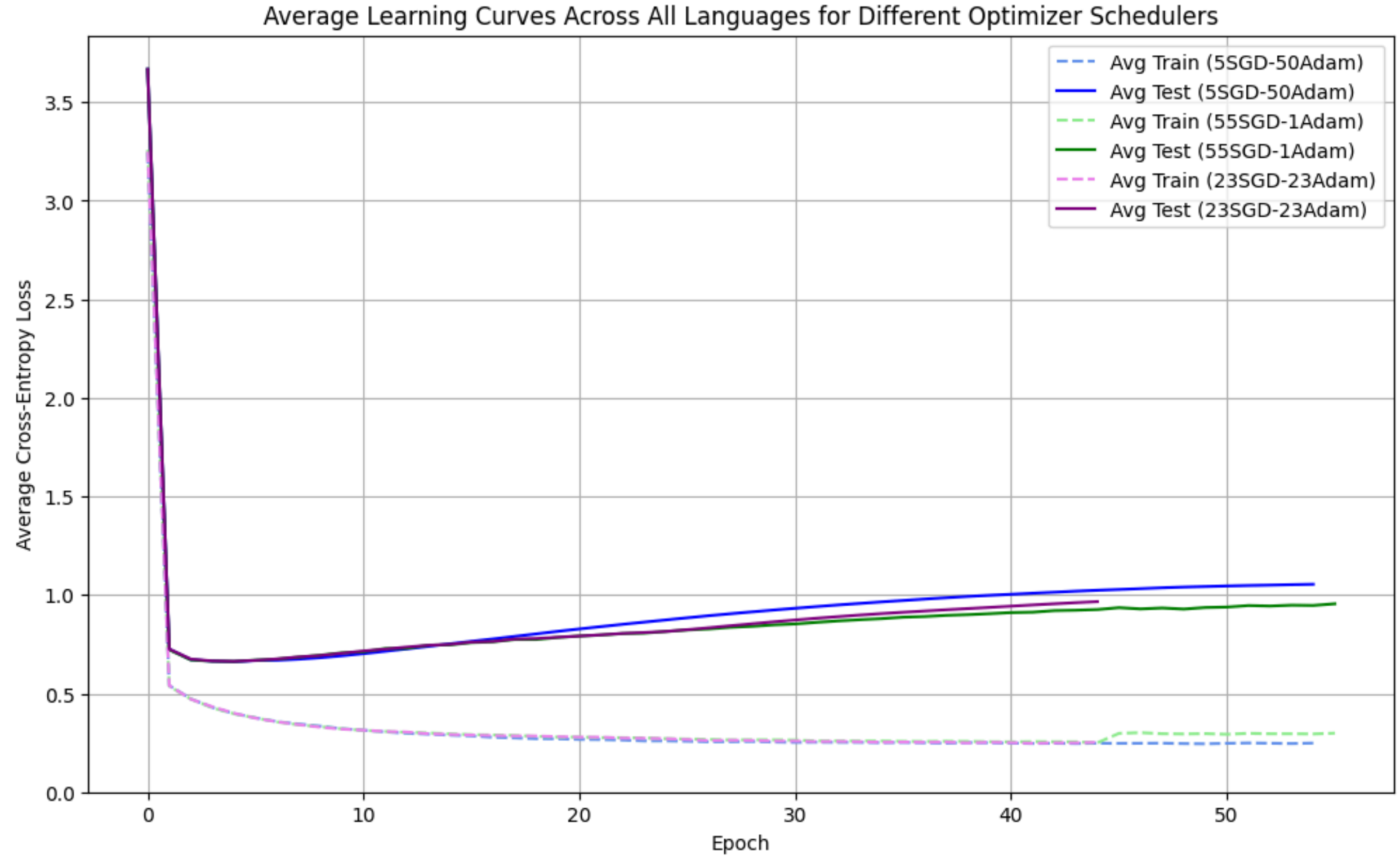

**Figure 4**. Training and test CE (train set size = 10, test set size = 1M) per epoch, averaged across Y&P's 56 formal languages, for different optimizer schedules.

## Experiment 2: evaluation metric

### Experimental setting

In this experiment, we focus on specific formal languages to demonstrate that contrary to M&G's claim, their meta-trained model does not match Y&P's Bayesian model in generalization capabilities. M&G base their claim on the observation that both models achieve comparable F-scores, computed by checking how many of the 25 most frequent strings in the language are generated by the model, and conversely, how many of the 25 most frequent strings generated by the model are valid strings in the language. We show that this evaluation metric is misleading, and that evaluation on a more comprehensive test set reveals substantial performance gaps between the two models.

We examine two languages for which Y&P's model successfully learned the correct underlying distribution: AnBn and AnBnCn. The AnBn dataset used by both Y&P and M&G is drawn from a probabilistic context-free grammar (PCFG) that follows a geometric distribution: at each step, the derivation terminates with probability $p = \frac{1}{3}$ by applying $S \rightarrow ab$, otherwise the recursive rule $S \rightarrow aSb$ is applied. The AnBnCn dataset is constructed by sampling strings from AnBn and appending the appropriate number of Cs to each string, yielding a similar distribution.

To evaluate M&G's meta-trained model on these languages, we first train it on the relevant language using training sets of varying sizes, following M&G's own procedure (with the exception of the 100,000 training size, which is not included in M&G's experiments). We then evaluate the trained model's CE loss on an exhaustive test set comprising all strings in AnBn with $n=1,\ldots,1000$ (and analogously for AnBnCn). We compute a weighted average per-token CE, where each string's tokens are weighted by the string's probability under the true grammar. These weights account for essentially the entire distribution mass, with the omitted tail being negligibly small (below $10^{-176}$).

Since differences in CE loss can be difficult to interpret, and since both AnBn and AnBnCn crucially require the ability to count, we introduce an additional metric designed to directly probe the model's counting abilities. For each n from 1 to 1000, we check whether the model correctly predicts the timing of the EOS token — that is, whether it assigns maximum probability to the EOS token at the correct timestep, and only at that timestep. For AnBn, succeeding on this metric requires the model to count the number of B's and match it against the number of A's. This metric is deliberately charitable: it does not require the entire output distribution of the model to match the true distribution, only that the model prefer the correct next token over incorrect alternatives at key timesteps. Even a model that only aims at approximating the target should meet this standard. For reference, we also report F-scores computed using the same methodology as M&G.

**Results**

The findings are summarized in Table 1. Since Y&P's model learned the true distribution for both languages, it would succeed at predicting the EOS timing for any n, and its test CE score would likewise be optimal and computable directly from the true grammar. Optimal F-scores for Y&P's model are taken directly from their reported results.

When given sufficient training data, M&G's model does attain a perfect F-score, matching the performance of Y&P's model. However, as explained in the main text, this is not necessarily indicative of true generalization, since many of the 25 most frequent strings in the language are typically seen during training.

Importantly, the two models diverge considerably on the other metrics. Starting with CE, M&G's test scores are all higher than Y&P's, which already indicates weaker performance, though the magnitude of this gap is hard to assess intuitively. Inspecting the EOS-timing metric makes this divergence clearer: M&G's model exhibits serious limitations, getting only a small fraction of string lengths right. In fact, as Figure 5 demonstrates, the model first fails on a relatively small n, and continues to fail at predicting the EOS timing for every n thereafter, with the predicted EOS position drifting farther and farther away from its correct position.

Note that this failure is not due to the model architecture's inability to represent the correct solution: LSTMs are known to be capable of perfectly representing AnBn and AnBnCn. These results thus indicate that, contrary to what the F-score comparison might suggest, M&G's method fails to replicate the systematic generalization exhibited by Y&P's Bayesian model, even with a substantial amount of training data.

| **Language** | **Model** | **Training size** | **Max n in training** | **F-score** (25 most frequent strings) | **Test CE** | **Largest n with correct EOS timing** (max tested n=1,000) |
|---|---|---|---|---|---|---|
| **AnBn** | **Y&P (Optimal baseline)** | **1,000** | **16** | **1** | **0.2728** | ∞ |
| | **M&G** | 10 | 4 | 0.6844 | 0.3268 | 51 |
| | | 100 | 9 | 0.84 | 0.2905 | 39 |
| | | 1,000 | 16 | 1 | 0.3057 | 38 |
| | | 10,000 | 25 | 1 | 0.2962 | 42 |
| | | 100,000 | 37 | 1 | 0.2952 | 45 |
| **AnBnCn** | **Y&P (Optimal baseline)** | **10** | **8** | **1** | **0.1909** | ∞ |
| | **M&G** | 10 | 8 | 0.8324 | 0.2917 | 21 |
| | | 100 | 11 | 0.7111 | 0.2111 | 15 |
| | | 1,000 | 20 | 1 | 0.217 | 38 |
| | | 10,000 | 27 | 1 | 0.2144 | 66 |
| | | 100,000 | 30 | 1 | 0.2132 | 82 |

**Table 1.** M&G's model vs. Y&P's Bayesian model, evaluated on AnBn and AnBnCn. Given sufficient training data, M&G's model achieves perfect F-scores on the 25 most frequent strings, but falls short on metrics computed over a more exhaustive test set.

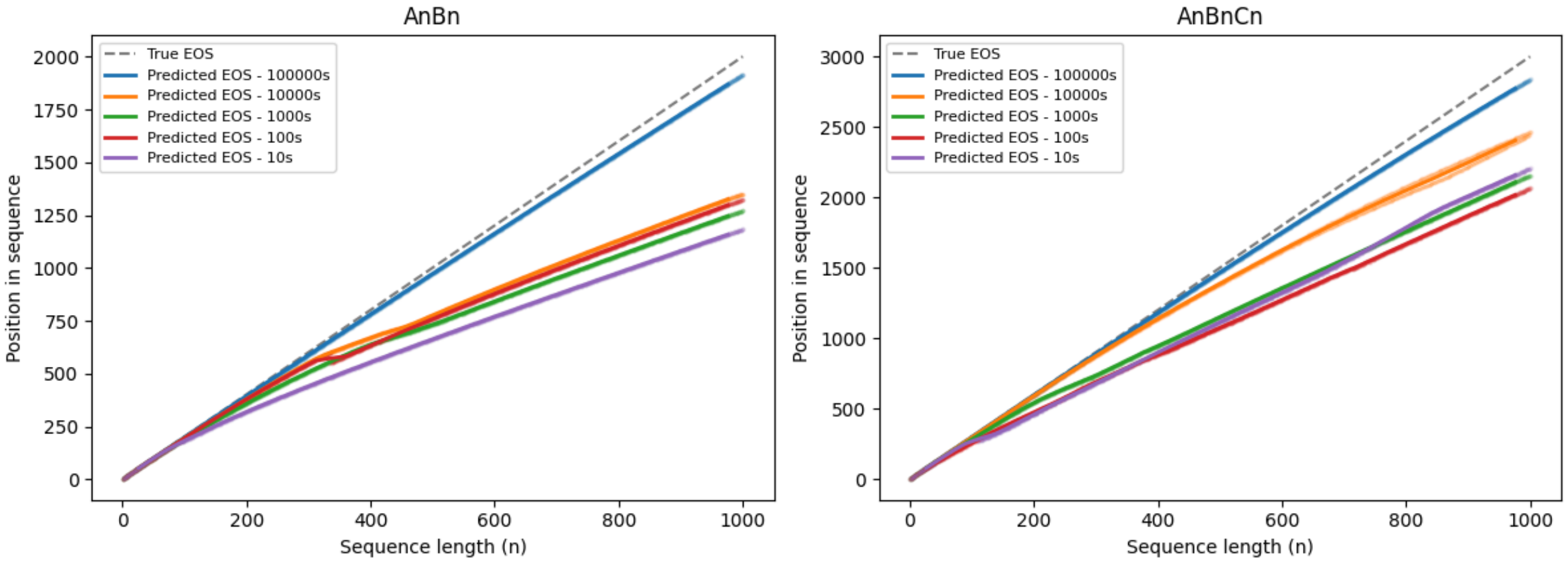

**Figure 5**. Predicted vs. ground-truth EOS position for AnBn and AnBnCn, across varying training set sizes. Even with more training data than M&G originally tested, their model still predicts the EOS increasingly earlier than its correct position as n grows.

## Appendix A: language definitions

A list of test formal languages with their definitions, taken from Y&P:

| | |
|---|---|
| An | Any number of As |
| AB | Sigma* over {A, B} |
| ABn | (AB)^n |
| AAA | A, AA, or AAA |
| AAAA | A, AA, AAA, or AAAA |
| AnBm | A^n B^m (n and m can be equal) |
| GoldenMean | Strings over {A, B} where no 2 A's ever appear in a row |
| Even | Strings over {A, B} where A's only appear in even-length groups |
| ApBAp | A+ B A+ |
| ApBApp | A+ (B A+)+ |
| AsBAsp | A* (B A*)+ |
| CountA2 | Strings over {A, B} where the number of A's is at least 2 |
| CountAEven | Strings over {A, B} where the number of A's is even |
| aABb | a Sigma+ b |
| AnBn | A^n B^n |
| Dyck | Balanced sequences of parentheses |

<table>
<tr><td>AnB2n</td><td>A^n B^(2n)</td></tr>
<tr><td>AnCBn</td><td>A^n C B^n</td></tr>
<tr><td>AnABn</td><td>A^n (AB)^n</td></tr>
<tr><td>ABnABAn</td><td>(AB)^n (ABA)^n</td></tr>
<tr><td>AnBmCn</td><td>A^n B^m C^n</td></tr>
<tr><td>AnBmA2n</td><td>A^n B^m A^(2n)</td></tr>
<tr><td>AnBnC2n</td><td>A^n B^n C^(2n)</td></tr>
<tr><td>AnBmCm</td><td>A^n B^m C^m</td></tr>
<tr><td>AnBmCnpm</td><td>A^n B^m C^(n+m)</td></tr>
<tr><td>AnBmCnm</td><td>A^n B^m C^(n*m)</td></tr>
<tr><td>AnBk</td><td>A^n B^(n+m)</td></tr>
<tr><td>AnBmCmAn</td><td>A^n B^m C^m A^n</td></tr>
<tr><td>AnB2nC3n</td><td>A^n B^2n C^3n</td></tr>
<tr><td>AnBnp1Cnp2</td><td>A^n B^(n+1) C^(n+2)</td></tr>
<tr><td>AnUBn</td><td>A^n | B^n</td></tr>
<tr><td>AnUAnBn</td><td>A^n | (A^n B^n)</td></tr>
<tr><td>ABnUBAn</td><td>(AB)^n | (BA)^n</td></tr>
<tr><td>XX</td><td>XX (two copies of the same string)</td></tr>
</table>

| XXX | XXX |
|---|---|
| XY | XY: X != Y (2 non-empty strings that are not identical; could include BBBB = B + BBB) |
| XXR | X X^R (even-length palindromes) |
| XXI | X X^I (where X^I is the inverse of X - replace every A with B and vice-versa) |
| XXRI | X (X^R)^I |
| An2 | A^(n^2) |
| AnBmCnDm | A^n B^m C^n D^m |
| AnBmAnBm | A^n B^m A^n B^m |
| AnBmAnBmCCC | A^n B^m A^n B^m CCC |
| AnBnCn | A^n B^n C^n |
| AnBnCnDn | A^n B^n C^n D^n |
| AnBnCnDnEn | A^n B^n C^n D^n E^n |
| A2en | A^(2^n) |
| ABnen | (AB)^(n^2) |
| Count | Any prefix of [b, bb, bbb, bbbb, ...], with a's separating the elements of the prefix |
| ChineseNumeral | Several groups of b's joined with a single a separating each; each group of b's must be shorter than the previous one |

<table>
<tr><td>ABAnBn</td><td>Sigma+ $A^n B^n$ (except it seems to be missing some short ones, like aab)</td></tr>
<tr><td>ABaaaAB</td><td>Sigma+ AAA Sigma+</td></tr>
<tr><td>Unequal</td><td>Strings over {A, B} with an unequal number of A's and B's</td></tr>
<tr><td>Bach2</td><td>Strings over {A, B} with an equal number of A's and B's</td></tr>
<tr><td>Bach3</td><td>Strings over {A, B, C} with an equal number of A's, B's, and C's</td></tr>
<tr><td>WeW</td><td>$X^{|X|}$ (that is, a string X repeated |X| times, where |X| is the length of X)</td></tr>
</table>